\documentclass{article}

\usepackage[preprint,nonatbib]{neurips_2024}

\usepackage[numbers]{natbib}
\usepackage[utf8]{inputenc} 
\usepackage[T1]{fontenc}    
\usepackage{hyperref}       
\usepackage{url}            
\usepackage{booktabs}       
\usepackage{amsfonts}       
\usepackage{nicefrac}       
\usepackage{microtype}      
\usepackage{listings}
\usepackage{csvsimple}
\usepackage{stackengine}
\usepackage{graphicx}
\usepackage{pdflscape}
\usepackage{changepage}

\usepackage{caption}    

\usepackage{multirow}
\usepackage[table,xcdraw]{xcolor}

\usepackage{caption}
\usepackage{enumitem}

\title{Introducing Milabench:\\Benchmarking Accelerators for AI}

\author{%
   Pierre Delaunay
   \And
   Xavier Bouthillier
   \And
   Olivier Breuleux
   \And
   Satya Ortiz-Gagné
   \And
   Olexa Bilaniuk
   \And
   Fabrice Normandin
   \And
   Arnaud Bergeron
   \And
   Bruno Carrez
   \And
   Guillaume Alain
   \And
   Soline Blanc
   \And
   Frédéric Osterrath
   \And
   Joseph Viviano
   \And
   Roger Creus-Castanyer
   \And
   Darshan Patil
   \And
   Rabiul Awal
   \And
   Le Zhang
   \AND
   \vspace{-1em}
   \\
   Mila \\
   \texttt{<first name>.<last name>@mila.quebec} \\
}

\begin{document}

\maketitle

\begin{abstract}
AI workloads, particularly those driven by deep learning, are introducing novel
usage patterns to high-performance computing (HPC) systems that are not
comprehensively captured by standard HPC benchmarks. As one of the largest
academic research centers dedicated to deep learning, Mila identified the need
to develop a custom benchmarking suite to address the diverse requirements of
its community, which consists of over 1,000 researchers. This report introduces
Milabench, the resulting benchmarking suite. Its design was informed by an
extensive literature review encompassing 867 papers, as well as surveys
conducted with Mila researchers. This rigorous process led to the selection of
26 primary benchmarks tailored for procurement evaluations, alongside 16
optional benchmarks for in-depth analysis. We detail the design methodology, the
structure of the benchmarking suite, and provide performance evaluations using
GPUs from NVIDIA, AMD, and Intel. The Milabench suite is open source and can be
accessed at github.com/mila-iqia/milabench.
\end{abstract}

\section{Introduction}

%
%
Recent breakthroughs in AI have been driven by a combination of factors,
including increased computing power, the availability of large datasets, and
advances in deep learning research.
Graphics Processing Units (GPUs) have become a cornerstone of modern AI
pipelines, with neural network architectures having co-evolved to leverage the
high parallelization capacity of GPUs.
Given that hardware and energy costs in HPC systems optimized for AI workloads
are predominantly driven by GPUs, selecting the right GPU is critical for
achieving optimal system performance within budget constraints.

%
AI workloads are inherently diverse, presenting unique challenges that are
difficult to replicate accurately using simple kernels or synthetic benchmarks.
To address this, Mila began developing an internal benchmarking suite
in 2019 to support its procurement process for an internal cluster dedicated to
its AI researchers. Over time, this benchmarking suite, Milabench, has matured
into a robust solution, which we present in this report.

The latest version of Milabench draws upon a comprehensive literature review of
867 papers published by Mila researchers in 2023. It encompasses a broad
spectrum of AI research domains, model architectures, and scaling pipelines, 
including 26 main benchmarks for procurement evaluations and 16
additional benchmarks for deeper analysis.

Milabench is designed and implemented based on 3 principal objectives:
\begin{enumerate}[leftmargin=0em,label=\textbf{\arabic*}.]
    \item \textbf{Simple to use:} To streamline the procurement process.
        \begin{description}[leftmargin=9em, style=nextline, align=right, font=\normalfont\bfseries\itshape\space]
        \item[Fast] Short test durations enable rapid experimentations.
        \item[Modular] A uniform interface across benchmarks simplifies both execution and contributions.%
    \end{description}
    \vspace{0.5em}
    \item \textbf{Representative:} To ensure the benchmarks reflect real-world usage.
    \begin{description}[leftmargin=9em, style=nextline, align=right, font=\normalfont\bfseries\itshape\space]
        \item[Diverse] Covers a wide range of research topics and models studied at Mila.
        \item[End to end] Addresses bottlenecks beyond GPU performance, reflecting the full pipeline.
        \item[Closed] Avoids vendor-specific optimizations that researchers are unlikely to use.
        \item[Dependencies] Incorporates libraries commonly used by researchers to validate compatibility.
    \end{description}
    \vspace{0.5em}
    \item \textbf{Unbiased:} To promote diversity in hardware and software solutions.
    \begin{description}[leftmargin=9em, style=nextline, align=right, font=\normalfont\bfseries\itshape\space]
        \item[Thorough Testing] Benchmarks are validated across multiple hardware vendors.
        \item[Generic Pipelines] Favors widely adopted alternatives over vendor-contributed pipelines to ensure generality.
    \end{description}
\end{enumerate}

We will describe in Section~\ref{sec:design} the design strategies
used to select the benchmarks in the latest version of Milabench, along
with a full list of the benchmarks included.
A brief overview of the implementation of the suite will be 
covered in Section~\ref{sec:tool}, before showcasing results
in Section~\ref{sec:results}. We will conclude by providing
a comparison with other benchmarks in Section~\ref{sec:related-work},
and discuss future directions in Section~\ref{sec:future}.

\section{Suite Design}

\label{sec:design}

The benchmarks included in Milabench are carefully selected and balanced to
reflect the breadth of research conducted at Mila. With a community of over
1,000 researchers, professors, and graduate students, the institute spans a
diverse range of research topics. The selection process is informed by surveys
and an extensive literature review of 867 papers published by Mila members in
2023.

This section begins with a detailed presentation of the literature review
(\ref{sec:lit-review}), followed by a discussion of the survey results
(\ref{sec:survey}). Finally, we outline the selection strategy derived
from these analyses (\ref{sec:choices}). 

\subsection{Literature Review}
\label{sec:lit-review}

The primary source for benchmark selection is the comprehensive literature
review. A total of 867 papers published by Mila researchers in 2023 were
collected using Paperoni, an open-source utility
developed at Mila, and converted to raw text using pdftotext~\cite{poppler2024}.
Annotation of the papers was performed using GPT-4o, extracting key information
such as research domains, model descriptions, and dataset details. An example
of a response from GPT-4o is provided in Appendix~\ref{app:gpt-4o}.

\subsubsection*{Evaluation}

To validate the accuracy of GPT-4o and assess the utility of the resulting
statistics, we manually annotated 110 papers published prior to 2024. To ensure
a broad distribution of research topics, a pool of 750 papers published between
2019 and 2024 was initially categorized using a naive approach: literal keyword
matching of 11 predefined domains within their abstracts\footnote{Audio,
    Computer Vision,
    Generative Models,
    Graph Neural Network,
    Medical,
    Multimodal,
    Natural Language Processing,
    Neuroscience,
    Recommendation System,
    Reinforcement Learning,
    Robotics,}. 
From each category, 10 papers were randomly sampled, resulting in a balanced set of 110 papers.
These 110 papers were processed twice by GPT-4o, yielding 220 responses. Both
responses for each paper, along with the corresponding PDF, were manually
reviewed, corrected, and merged to create a high-quality annotation set.

To assist in validating GPT-4o's predictions, the JSON template provided to the
model included fields for justifications and quotes supporting each value. This
strategy was applied to both research domains and model names to enhance
transparency and interpretability.
The validation set was subsequently used to evaluate GPT-4o's performance via a
Multi-Label Confusion Matrix (MLCM)~\cite{mlcm}, focusing on the accuracy of
predictions for research domains and model types.

Predictions often exhibited significant variability across papers, with frequent
overlaps between domains and model architectures. Additionally, many papers
described numerous models, which GPT-4o rarely captured comprehensively. To
address these challenges, we grouped models into broader architecture families
rather than requiring precise identification.
A hierarchical taxonomy of domains and models was manually created to aggregate
statistics at a higher level of abstraction. For instance, specific models like
Llama were mapped to broader categories such as Transformer.
Table~\ref{tab:lit-review-mlcm} presents quality measures for GPT-4o’s
predictions compared to manual annotations, aggregated using this hierarchy.
In this evaluation, predictions were mapped to the hierarchy's top level. For a
given paper, a top-level category was assigned a count of 1 if there was at
least one match and 0 otherwise. This aggregation strategy provided a more
comprehensive overview and improved both recall and precision by over 20\%.
The hierarchy included numerous top-level categories beyond those shown in
Table~\ref{tab:lit-review-mlcm}. Less descriptive categories (e.g., Deep
Learning Optimization or Neural Network) and those with sparse occurrences were
consolidated into the category \texttt{Others}.

\subsubsection*{Results}

We observe recall rates ranging from 84\% to 98\% and precisions ranging from
86\% to 100\% for domains, 
with averages of 94\% for both metrics. For model architectures, 
recalls ranged from 50\% to 100\% and precisions ranged from 84\% to 100\%, with
an average of 75\% and 96\% respectively.

\begin{table}[b]
    \caption{MLCM tables for the domains and model with other domains or model architectures identified by GPT-4o. Additional column and row \textbf{NPL} and \textbf{NTL} stands for No Predicted Label and No True Label respectively. \\ }

    \resizebox{\textwidth}{!}{
    \begin{tabular}{llp{0.90em}p{0.90em}p{0.90em}p{0.90em}p{0.90em}p{0.90em}||lp{0.90em}p{0.90em}p{0.90em}p{0.90em}p{0.90em}p{0.90em}p{0.90em}p{0.90em}p{0.90em}}
    \multicolumn{8}{c||}{\textbf{Domains}} & \multicolumn{10}{c}{\textbf{Model Types}}
    \\
    \hline \hline
    & & \multicolumn{6}{c||}{\textbf{Predicted}} & & \multicolumn{9}{c}{\textbf{Predicted}}
    \\

    & \textbf{} & \rotatebox{90}{\textbf{Computer Vision}} & \rotatebox{90}{\textbf{Graphs}} & \rotatebox{90}{\textbf{Natural Language P.}} & \rotatebox{90}{\textbf{Reinforcement L.}} & \rotatebox{90}{\textbf{Others}} & \rotatebox{90}{\textbf{NPL}} &
    \textbf{} & \rotatebox{90}{\textbf{CNN}} & \rotatebox{90}{\textbf{Diffusion model}} & \rotatebox{90}{\textbf{Gflow nets}} & \rotatebox{90}{\textbf{GNN}} & \rotatebox{90}{\textbf{MLP}} & \rotatebox{90}{\textbf{RNN}} & \rotatebox{90}{\textbf{Transformer}} & \rotatebox{90}{\textbf{N/A}} & \rotatebox{90}{\textbf{NPL}}
    \\

    \multirow{9}{0em}{\rotatebox{90}{\textbf{True Labels}}}
    & \textbf{Computer Vision}     & \textbf{15} & 0 & 0 & 0 & 1 & 2 &
    \textbf{CNN}             & \textbf{23} & 0 & 0 & 0 & 0 & 0 & 0 & 1 & 6 \\
    & \textbf{Graphs}              & 0 & \textbf{6}  & 0 & 0 & 2 & 2 &
    \textbf{Diffusion model} & 0 & \textbf{4}  & 0 & 0 & 0 & 0 & 0 & 0 & 1 \\
    & \textbf{Natural Language P.} & 0 & 0 & \textbf{17} & 0 & 0 & 4 &
    \textbf{Gflow nets}      & 0 & 0 & \textbf{5}  & 0 & 0 & 0 & 0 & 0 & 0 \\
    & \textbf{Reinforcement L.}    & 0 & 0 & 0 & \textbf{15} & 3 & 3 &
    \textbf{GNN}             & 0 & 0 & 0 & \textbf{10} & 0 & 0 & 0 & 0 & 4 \\
    & \textbf{Others}              & 1 & 1 & 1 & 0 & \textbf{92} & 7 &
    \textbf{MLP}             & 0 & 0 & 0 & 0 & \textbf{2}  & 0 & 0 & 0 & 1 \\
    & \textbf{NTL}                 & 0 & 0 & 0 & 0 & 0 & 0 &
    \textbf{RNN}             & 0 & 0 & 0 & 0 & 0 & \textbf{3}  & 0 & 0 & 3 \\
    &                              &   &   &   &   &   &   &
    \textbf{Transformer}     & 0 & 0 & 0 & 0 & 0 & 0 & \textbf{38} & 2 & 6 \\
    &                              &   &   &   &   &   &   &
    \textbf{N/A}             & 0 & 0 & 0 & 1 & 0 & 0 & 0 & \textbf{48} & 18 \\
    &                              &   &   &   &   &   &   &
    \textbf{NTL}             & 1 & 0 & 0 & 0 & 0 & 0 & 1 & 6 & 0 \\

    \hline \hline
    & \textbf{Precision (\%)}      & 94 & 86 & 94 & 100 & 94 & &
                             & 96 & 100 & 100 & 91 & 100 & 100 & 97 & 84 & \\
    & \textbf{Recall (\%)}         & 98 & 97 & 97 & 96  & 84 & &
                             & 77 & 80  & 100 & 71 & 67  & 50  & 83 & 72 &
    \end{tabular}
    }
    \label{tab:lit-review-mlcm}
\end{table}

These results served as targets to balance the benchmarks in Milabench, ensuring
they represent the relative proportions of research domains and model
architectures used by Mila researchers. Figure~\ref{fig:design-tables}
illustrates these proportions compared to the benchmarks included in the suite.

One notable finding from the literature review is the significant proportion of
research focused on or involving Reinforcement Learning (RL), a domain
previously absent from earlier versions of Milabench. While Computer Vision and
Natural Language Processing (NLP) were expected to dominate as the most
prominent research domains, the results reveal that NLP and RL hold similar
levels of importance at Mila.
An additional insight is the nuanced role of Computer Vision in research. Only
about 19\% of papers were explicitly categorized as focusing on CV by GPT-4o.
However, an additional 18\% of papers used vision datasets without being
explicitly categorized under CV, often being labeled as \texttt{Deep Learning
Neural Network} or \texttt{Deep Learning Optimization}. This brings the total
proportion of papers involving CV tasks to 37\%. Of these, roughly half focus on
CV-specific studies, while the other half employ CV tasks as a general
experimental platform.
A similar pattern, though less pronounced, was observed for NLP. Approximately
4.4\% of papers utilized NLP datasets without being categorized under NLP.
Conversely, this phenomenon is negligible in the case of RL, where papers
generally align directly with their core research domain.

The proportion targets for research \emph{domains} used in the design of
benchmarks, as shown in Figure~\ref{fig:design-tables} (blue bars), are derived
from a combination of domain categorizations and dataset domain categorizations.
In contrast, the proportion targets for \emph{model architectures} are based
directly on the statistics obtained from GPT-4o.

\subsection{Surveys}
\label{sec:survey}

Surveys were conducted to address gaps in the literature review and to account
for emerging trends. These surveys primarily aimed to gather statistics on the
most commonly used libraries and to identify potential benchmark pipeline
candidates.

\subsubsection*{Libraries}

Only a small fraction of papers explicitly cite or mention the libraries used
for the experiments presented. As a result, we relied on internal surveys to
identify the most widely used libraries for inclusion in Milabench.

PyTorch is by far the most commonly used library, with 96\% of Mila researchers
adopting it. In contrast, TensorFlow is used by only 5\% of researchers, all of
whom also use PyTorch. The adoption of JAX has been steadily increasing, with
26\% of researchers now using it. Among these, 2.6\% use JAX exclusively, while
the remaining 23.4\% use both JAX and PyTorch. Given these usage patterns, we
decided to include tests using only PyTorch and JAX. JAX is primarily utilized
for reinforcement learning research, a trend that is reflected in the suite.

Both NLP and Computer Vision researchers have increasingly adopted Hugging Face, with approximately 33\% of researchers using it. Additionally, the generic framework PyTorch Lightning is used by 26\% of Mila researchers.



\subsubsection*{Benchmark pipeline candidates}

Professors were surveyed to suggest new pipelines for Milabench. 
However, the responses were highly biased, underscoring the need for a more
reliable method to gather statistics on the domains to be covered. A comparison
of survey results with our literature review revealed notable discrepancies. For
instance, 60\% of the suggested pipelines focused on Computer Vision, while only
37\% of papers in the literature review were dedicated to this domain.
Similarly, 15\% of suggested pipelines were for NLP, compared to 26\% in the
literature, 20\% for RL versus 26\%, and 5\% for Graphs
versus 15\%.

Given these biases, we relied on the expertise of the professors to suggest more
pipelines than necessary but ultimately used the literature review to make the
final selection.
A total of 25 pipelines were suggested, from which 10 were chosen. We reused 6
pipelines from the previous version of Milabench and added 3 additional
pipelines to address obvious gaps.

\subsection{Selection of benchmarks}
\label{sec:choices}

Coverage targets for the benchmarking suite were defined across several
dimensions, including research domains, model architectures, model sizes,
training parallelization methods, and libraries. These targets are not mutually
exclusive, meaning that some benchmarks can cover multiple areas simultaneously,
allowing for broader coverage with fewer tests.

The pipelines were selected based on the popularity of models and datasets, as
well as the availability of mature open-source implementations that are
well-supported by the research community. We rely on full implementations from
third-party libraries and make only minimal modifications to integrate them into
Milabench. Additionally, we avoid using models or frameworks contributed by
vendors to minimize potential biases in favor of those vendors.

The main benchmarking suite consists of 26 benchmarks, encompassing 19 different
model architectures. For certain model architectures, there are multiple
variations in execution pipelines to reflect the diverse usage patterns of
researchers. For example, benchmarks using Llama include variations for
pre-training with different parallelization budgets, fine-tuning with LoRA also
including different parallelization budgets, and an inference tasks, totaling 6
distinct benchmarks. The selection of main benchmarks and their corresponding
targets is illustrated in Figure~\ref{fig:design-tables}. A detailed list of the
main benchmarks can be found in Table~\ref{tab:main-benchmarks}.

In addition to the 26 main benchmarks, 16 optional benchmarks are available for
further analysis. These include variations in precision formats (fp16, bf16,
tf32, and fp32) for basic matrix multiplication, transformer training 
(BERT), and CNN training (ConvNext). The benchmark \texttt{resnet50} is
sensitive to I/O and CPU performance, so we included an additional version
without I/O to measure the impact of potential bottlenecks caused by low
bandwidth or insufficient CPU resources.

Some main benchmarks are only included as multi-GPU tasks (e.g.,
\texttt{dinov2-giant-gpus}) because training these models on a single GPU would
be too slow for practical use. For these, we provide single-GPU versions as
optional benchmarks, allowing for scaling measurements from single-GPU to
multi-GPU setups within a single node, and from single-node to multi-node
configurations (e.g., \texttt{diffusion-nodes}).
While \texttt{rlhf-single} would also be more efficiently trained using multiple
GPUs, we include the single-GPU version as the main benchmark and the multi-GPU
version as an optional benchmark to avoid further biasing the suite toward
single-node training.

A detailed list of the optional benchmarks is
provided in Table~\ref{tab:additional-benchmarks}. The optional benchmarks are
not represented in Figure~\ref{fig:design-tables}, as they do not contribute
to achieving the coverage targets.

\section{Implementation}
\label{sec:tool}

Milabench is composed of four main components, milabench, benchmate, voir and torchcompat.

\begin{itemize}
    \item \textbf{milabench}: manage the benchmarking process, from its configuration to its execution,
    including the installation of dependencies and the downloads of datasets.
    
    \item \textbf{benchmate}: common utilities that can be reused between benchmarks,
    such as custom TimedIterators or custom monitors to log metrics.
    
    \item \textbf{voir}: instrumentation tool to retrieve and log metrics.
    It includes GPU monitors for different vendors and custom multiprocessing tools to help log metrics.

    \item \textbf{torchcompat}: compatibility layer to smooth out differences between vendors.
    It implements a superset of \texttt{torch.cuda} to enable all vendors to run seamlessly with no code change.
\end{itemize}

Each benchmark in Milabench is implemented as an independent "package." These
packages are designed to be isolated from one another, allowing each to install
its dependencies within its own virtual environment to prevent dependency
conflicts.

However, many benchmarks share common dependencies, and to optimize resource
usage, we have consolidated these shared dependencies into a single virtual
environment. This approach not only reduces the overall size of Milabench but
also simplifies execution. In practice, we strive to use a single environment
across benchmarks, which simplifies result analysis and helps keep Milabench
installation lightweight.

Milabench includes helper commands to manage dependency versions easily,
allowing them to be pinned to the latest compatible releases.

The execution of Milabench is organized into three steps. The first two steps
comprise an initial setup process, which can be skipped after they have been
successfully run once.

\begin{itemize}
    \item \textbf{install}: Create virtual environment for each benchmarks and install their dependencies

    \item \textbf{prepare}: Download checkpoints and datasets that will be required to run the benchmark suite
    
    \item \textbf{run}: Execute the benchmark suite

    \item \textbf{report} [Optional]: Generate an aggregate report of all the runs
\end{itemize}

The benchmark suite is fully defined by a YAML configuration file, which
specifies the benchmarks to be executed and their execution parameters. Users
can create customized benchmark suites by selecting from the pool of available
benchmarks.

Additionally, we provide a set of Docker images to simplify benchmark execution.
These images come with pre-installed dependencies to ensure better
reproducibility of results.

\subsection{Benchmarking Methodology}
\label{sec:methodology}

To provide metrics that closely reflect real-world scenarios, the benchmarks in
Milabench are designed to mirror the experimental pipelines commonly used by
researchers. This approach requires not only selecting the most popular models
and libraries but also ensuring that the code closely follows typical research
workflows, with minimal adaptations for vendor support and without introducing
additional optimizations. This design philosophy also simplifies the performance
measurement process.

By relying on third-party libraries that are well-supported by the community, we
focus on measuring system throughput rather than optimizing for convergence time
in training objectives. As a result, the 42 benchmarks in Milabench can be
executed in under 2 hours\footnote{assuming datasets and checkpoints have been
pre-downloaded} on the latest available GPUs.

Each benchmark is implemented as a short execution to gather 30 to 60
measurements of processing time for units of work. The nature of these units of
work varies depending on the benchmark and can be defined in terms of the number
of samples, tokens, nodes, or steps within episodes.

Where possible, datasets are generated randomly to minimize reliance on external
data sources. However, in some cases, subsets of standard datasets are used when
the characteristics of the data affect processing time. For example, in LLM
tasks, pure random noise would result in the LLM consistently producing
maximum-length sequences. The randomly generated datasets are sized to match
popular real-world datasets; for instance, FakeImageNet consists of images sized
at (384, 384), based on the average image sizes in ImageNet~\cite{imagenet},
and FakeVideos consists of 30 FPS, 10-second videos of size (640, 480), modeled after
the Kinetics dataset~\cite{kay2017kinetics}.

This approach is also applied to model initialization, particularly in inference
mode. Using a pre-trained model ensures performance metrics are aligned with
real-world scenarios, as using an untrained model would significantly skew
results.

Pipelines that typically incorporate online data augmentation are implemented as
such, with the augmentation process affecting the overall performance
measurement. Otherwise, data preprocessing is handled during the
preparation phase and is not included in the performance evaluation.

Tasks may be executed on three different scales: 1) single-gpu, 2) single-node
multi-gpu or 3) multi-node multi-gpu. In all cases, all available GPUs are used
to measure full-node performance. For single-GPU tasks, the benchmark is
launched separately on each GPU, and performance is averaged across GPUs. This
approach helps identify issues such as insufficient CPU resources, which may
hinder GPU performance. For multi-GPU tasks, whether on a single node or across
multiple nodes, performance is measured across the GPUs without any normalization.

Scores are aggregated across all main benchmarks using a weighted geometric
mean, implemented via the logsumexp function for improved numerical
stability. Benchmarks performances are multiplied by their success rate,
then increased by 1 to avoid $\log 0$. This gives the following
equation:

\begin{equation}
    \exp\left(\frac{\sum_i w_i\log{(p_i*s_i+1)}}{\sum w_i}\right)
\end{equation}

Where $i$ is the benchmarks index, $w_i$ is a weight, $p_i$ is a performance
measure and $s_i$ is a success rate.

\section{Results from Milabench}
\label{sec:results}

We executed Milabench across four different node pairs, each equipped with
distinct GPU models: Nvidia A100 SXM4 80GB, Nvidia H100 SXM5 80GB, AMD MI300X
192GB, and Intel Gaudi2 96GB. The performance results from the A100 were used as
the baseline for comparison with the H100, MI300X, and Gaudi2 GPUs. For further
details on the node specifications, please refer to Table~\ref{tab:node-specs}.

\subsection{Optional Benchmarks}
\label{sec:synth-results}

\begin{figure}[ht]
    \centering
    \includegraphics[width=\textwidth]{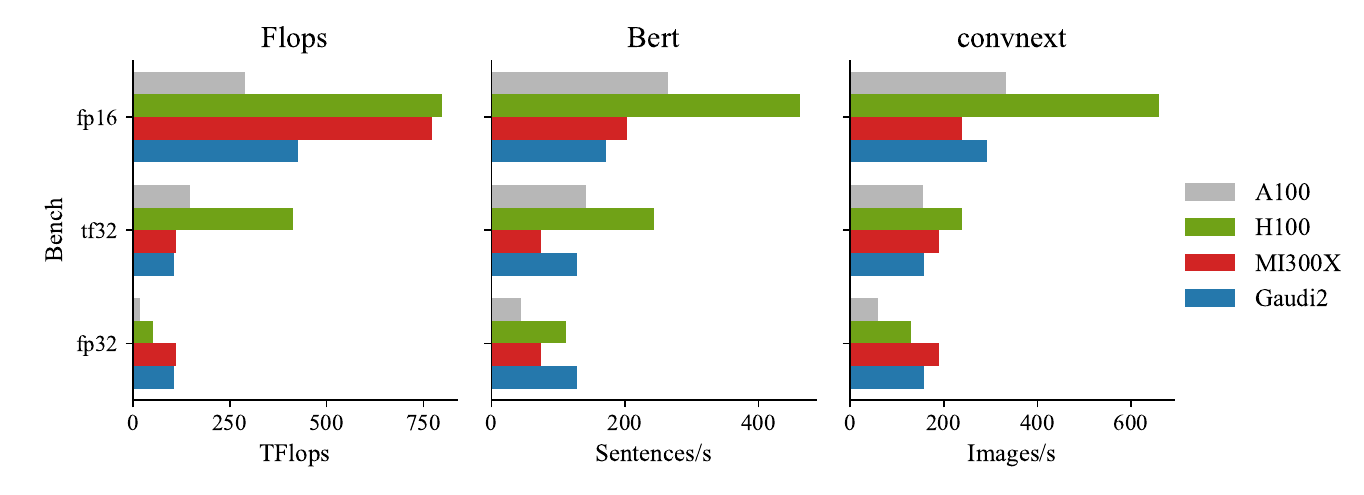}  
    \caption{Performance comparison between different data types\protect\footnotemark}
    \label{fig:flops}
\end{figure}

Before delving into the main results from Milabench, we first examine the FLOP
performance for each vendor's GPUs (Figure~\ref{fig:flops}). The data reveals
distinct design priorities across vendors. Both the Gaudi2 and MI300X emphasize
float32 performance, whereas the H100 focuses on low-precision operations,
particularly with TF32.

In terms of low-precision performance, the H100 and MI300X appear to be on par,
while the Gaudi2 lags behind \footnote{It is worth noting that the Gaudi2 was
released in May 2022, while the H100 and MI300X were released in March 2023 and
December 2023, respectively.}.

In the synthetic FLOP benchmark, the standout feature of the H100 is its
performance with TF32. While the other vendors support TF32, they do not show a
significant performance improvement over the A100. The official MI300X
specification mentions TF32, but we did not observe any notable gains compared
to single-precision performance. This could be due to a lack of support in
PyTorch. The Gaudi2, on the other hand, may default to using TF32 in PyTorch
without an option to disable it, which could obscure its performance
characteristics.

Despite the strong performance of MI300X and Gaudi2 observed in the synthetic
FLOP benchmark, the results do not fully translate into real-world workloads,
such as BERT and ConvNext training (Figure~\ref{fig:flops}). The H100
outperforms both the MI300X and Gaudi2 in low-precision settings, even though it
was comparable to the MI300X in the synthetic low-precision (FP16) benchmark.
Moreover, in single-precision tasks, the H100 demonstrates competitive
performance with the MI300X and Gaudi2, despite the latter two GPUs achieving
significantly higher FLOP counts in the synthetic benchmark.

The full results for optional benchmarks are presented in Table~\ref{tab:optional-results}.

\footnotetext{Note that Gaudi2 might be using TF32 on the fp32 benchmark.}

\subsection{Main benchmarks}

\begin{figure}[ht]
    \centering
    \includegraphics[width=0.9\textwidth]{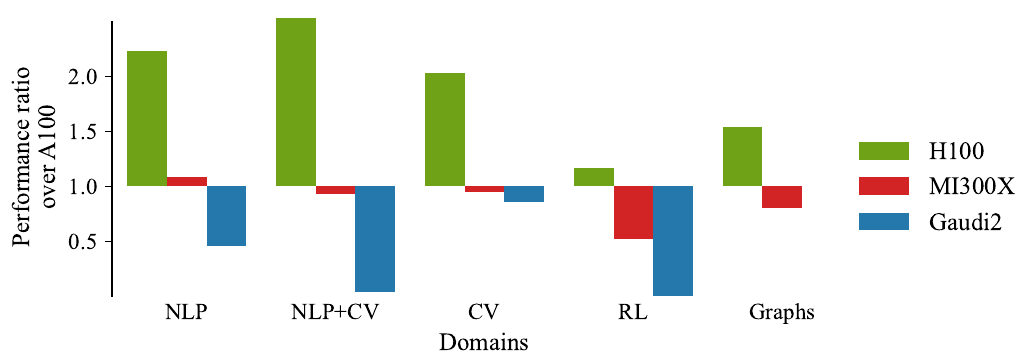}  
    \caption{Results Overview \protect\footnotemark}
    \label{fig:overview}
\end{figure}

Finally, as seen in the result overview in Figure~\ref{fig:overview}, the trends
observed for BERT and ConvNext generally extend across other domains as well.
Despite the MI300X matching the H100 in terms of low-precision FLOPS and
outperforming it in single-precision performance, the MI300X is consistently
outperformed in all benchmarks.

We attribute this performance disparity between FLOP counts and real-world
performance to differences in software stack maturity. CUDA has been the de
facto standard for AI and machine learning workloads for over a decade. During
this time, its compute kernels have been continually tuned and developed
alongside AI advancements, and it remains tightly integrated with the most
popular machine learning libraries. As the default platform, CUDA naturally
benefits from extensive testing and optimization. In contrast, newer platforms
need to invest considerable effort to catch up to years of innovation and
optimization inherent in CUDA.

AMD and Intel have taken different approaches to address this challenge. AMD
appears to have chosen a broad strategy, emulating CUDA and focusing on
achieving feature parity. This approach is evident in the wide test coverage,
with only one benchmark unsupported. Intel, on the other hand, has opted for a
more targeted approach, focusing on high-demand applications and developing
highly optimized, customized implementations. Milabench’s design favors the
first approach, as it aims to faithfully represent common experimental
pipelines, avoiding vendor-specific re-implementations. If a benchmark is
supported in Milabench, it indicates that there is a low barrier to entry for
working with the corresponding GPUs in that research domain.

\footnotetext{Note that Gaudi2 did not support any of the graphs benchmarks}

\begin{description}[style=nextline]
    \item[Natural Language Processing]
    NLP has gained significant popularity in recent years, with Large Language
    Models (LLMs) becoming a primary focus for all vendors. While the
    performance gap across GPUs in NLP is smaller compared to other domains,
    there is still noticeable variation. All GPUs perform well with full Llama
    pre-training, but fine-tuning with LoRA presents challenges for the MI300X
    and Gaudi2. We anticipate that major improvements will be seen in this area
    in future releases.

    \item[Computer Vision] 
    Computer Vision is the most supported domain, with all vendors performing
    well on vision models. Older models like ResNet50 are particularly
    well-optimized. Newer or less popular models, however, perform better on the H100.

    \item[Reinforcement Learning]
    Historically less compute-intensive than other domains, RL is now leveraging
    GPUs for environment execution using JAX, introducing new usage patterns for
    GPU benchmarks. There is yet major improvements to come in this field
    from every vendors. H100's averaged result is boosted mainly
    by the 2 tasks \texttt{recursiongfn} and \texttt{rlhf-single} that involves
    transformers, but makes its smallest gains of all benchmarks over A100 on
    the more canonical tasks.

    \item[Graphs]
    The Graph domain has been rapidly evolving, making support more challenging.
    Graphs typically rely on sparse operations, which are not supported by all
    vendors. Moreover, many custom CUDA kernels are built during installation,
    which complicates vendor support. Despite these challenges, the MI300X
    successfully ran all benchmarks, while Gaudi2 lacked support for compiling
    custom CUDA kernels into a compatible format.
 
\end{description}

The full results for main benchmarks are presented in Table~\ref{tab:main-results}.

The AI field is evolving quickly, and the results presented here are likely to
become outdated soon. New PyTorch versions often bring performance improvements
that range from small 5\% gains to more than 100\% on specific benchmarks.
These performance gains can be attributed to algorithmic improvements that apply
to all vendors but can also be attributed to improved integrations with some
vendor stacks.
Based on the discrepancies observed between synthetic and real-world benchmarks
(as discussed in Section~\ref{sec:synth-results}), we expect future software updates to
significantly impact GPU performance. This discrepancy seems to stem more from
differences in software stack support than hardware limitations. In short, stay
tuned for result updates.

\section{Related Work}
\label{sec:related-work}

High-Performance Computing (HPC) has a long-standing tradition of benchmarking,
with tools like LINPACK \cite{dongarra1979linpack} still being used to determine
the TOP500 list of the world’s most powerful computer systems. 
Although LINPACK offers valuable insights into the raw performance of a system,
it does not always capture the performance characteristics observed in
real-world applications.

In recent years, HPC has more widely adopted
procurement practices relying strongly on application-based benchmarks 
in addition to pure kernel benchmarks. This transition aims to provide a more accurate
assessment of system performance, aligned with the anticipated usage patterns of
new systems.
For example, the Jülich Supercomputer Center (JSC) has developed a
benchmark suite consisting of 16 application benchmarks and 7 synthetic
benchmarks \cite{herten2024application}. Of these, 3 are AI-focused. These
benchmarks include: \texttt{MMoCLIP}, A multi-modal text-image task using a ViT-L-14 model,
trained across 8 nodes for a total of 32 GPUs; 
\texttt{Megatron-LM}, A 175 billion-parameter language model trained on a
language modeling task, trained across 96 nodes for a total of 384 GPUs; and
\texttt{ResNet}, A ResNet50 model trained for image classification, executed on
10 nodes for a total of 40 GPUs. 
While \texttt{MMoCLIP} and \texttt{Megatron-LM} are excellent candidates to
benchmark large-scale AI training, the \texttt{ResNet} benchmark is based on a
model that may be too small to fully exploit the capabilities of modern GPUs in
large-scale training scenarios. The benchmark suite does not cover
important AI domains such as reinforcement learning (RL) and graph learning. This
omission is likely intentional however, as these domains are typically executed
on a single GPU or node, whereas the suite is specifically tailored for
large-scale training environments.

The benchmark requirements for general HPC system procurement must cover a broad
spectrum of research areas, which makes it challenging to allocate more than a
few tests specifically for AI. In contrast, for AI-focused benchmarking, MLPerf
\cite{mlperf}, developed and supported by MLCommons, serves as a key reference. Initially
focused on training benchmarks, MLPerf has since expanded to include benchmarks
for inference \cite{reddi2020mlperf} and for HPC systems \cite{farrell2021mlperf}.
Rather than prescribing specific
implementations, MLPerf provides guidelines and rules for implementing
pipelines, thereby allowing vendors the flexibility to introduce optimizations
and re-implementations. This approach fosters innovation by allowing significant
modifications while still ensuring the model quality meets the target standards.
However, this flexibility comes at the cost of potentially poor reproducibility
of performance results, the vendor-specific optimizations being trapped in
vendor implementations rather than contributed to popular libraries supported by
the community of researchers.
In contrast, Milabench emphasizes the use of precise, standardized
implementations to ensure robust reproducibility and greater generalizability of
performance across similar pipelines. Another notable distinction between MLPerf and
Milabench is the reporting methodology. MLPerf allows vendors to submit results
for individual benchmarks, without providing a unified framework for aggregating
these results into an overall performance measure. While this fragmented
approach may benefit vendors who wish to highlight specific strengths, it
creates a disjointed landscape of results, making it difficult to draw general
comparisons across different vendors. Milabench, on the other hand, adopts a
more holistic approach, striving for comprehensive coverage across benchmarks.
This ensures that researchers at Mila are not left behind when acquiring
hardware that may not fully support their research tools. Consequently,
Milabench aims to acquire complete sets of results, where any missing benchmark
negatively impacts the overall score, ensuring that the system's performance is
evaluated in a manner that is both comprehensive and consistent.

\section{Future Directions for Milabench}
\label{sec:future}

We aim to update Milabench annually to ensure that its benchmarking pipelines
reflect current research advancements and incorporate support for newer hardware
as it becomes available. However, obtaining hardware from multiple vendors for
testing purposes remains a significant challenge, which has limited our ability
to maintain continuous integration across a diverse set of hardware platforms.
To address this, we are actively working to expand support for a broader range
of vendors and to develop diagnostic tools that can assist in identifying
potential issues and optimization opportunities for these systems.

Currently, the literature review process that underpins Milabench still requires
substantial manual intervention, particularly in curating domains and model
architectures into a structured hierarchical taxonomy. To scale the review process and
enable it to handle a larger corpus of publications, we are exploring automation
techniques that would allow the literature review to be updated annually,
streamlining the incorporation of new research.

Additionally, the current classification of model architectures in Milabench
does not fully account for the diversity of operations within individual models.
For example, many models include dense layers at the output stage, which are
structurally similar to Multi-Layer Perceptrons (MLPs). Autoencoders, too, are
often integrated with other models or serve as components within larger
architectures. We believe that rethinking the categorization of model
architectures to focus on individual components or "architecture parts" will
provide a more accurate representation of the computational patterns prevalent
in AI pipelines.

Furthermore, there has been limited attention given to variations in input sizes
beyond the selection of popular dataset formats. Future benchmarking targets
should incorporate not only the design of input/output sizes but also include
optional benchmarks that measure the impact of varying input sizes across
different GPUs. This would help evaluate the scalability and performance
sensitivity of models to input size changes, which is crucial for optimizing AI
workloads on modern hardware.

Finally, as energy consumption becomes an increasingly important concern with
the proliferation of densely packed HPC systems, we recognize the need to
incorporate energy efficiency into the global performance metric of Milabench.
Future versions of the benchmark suite should account for energy costs in
addition to raw performance, providing a more comprehensive assessment of system
efficiency and sustainability in high-performance AI and HPC environments.

\section{Conclusion}
\label{sec:conclusion}

Milabench offers a comprehensive solution that allows vendors to quickly test
and experiment with different configurations across a wide range of models. At
the same time, it gives customers the flexibility to design custom benchmarks
tailored to their specific needs, enabling them to choose the hardware that best
aligns with their requirements.

Beyond providing insights into hardware performance, Milabench also evaluates
the maturity of the vendors' software stacks. This dual focus ensures that the
needs of research institutions and compute centers are thoroughly addressed.
Additionally, it provides vendors with valuable feedback on how to improve their
software stacks, highlighting areas with weak coverage, potential blind spots,
and suboptimal compute kernels. This helps vendors optimize both their hardware
and software solutions more effectively.

\nocite{*}

\bibliographystyle{plain}

\bibliography{references}

\begin{landscape}
    \begin{figure}[ht]
    \vspace{-3em}
    \includegraphics[width=1.7\textwidth]{img/paper\_design\_all-v1.4-no-dqn.png}
    \vspace{-25em}
    \captionsetup{indention=0cm,margin={1cm, 0.66\linewidth}}
    \caption{Design breakdown across 5 different major characteristics. 
    The beige columns represent the benchmarks included in the main suite. 
    The height of the cells represents their weight (ex: dimenet has a weight 
    of 2
    while ppo has a weight of 1). The blue bars are the targets we determined 
    based on our literature review and internal surveys. Each benchmark 
    appears at least once in all of these tables. The columns are not 
    mutually exclusive
    in all tables except for model sizes. To design Milabench's suite, 
    we selected benchmarks such that the proportions of the benchmarks in each
    column of these tables are as close as possible to the targets.}
    \label{fig:design-tables}
\end{figure}
\end{landscape}

\break

\begin{landscape}

\begin{table}[ht]
\bgroup
\def\arraystretch{1.2}
    \vspace{-4em}
\caption{Main benchmarks included in Milabench. Benchmarks with no description for parallelism are not parallelized and rather executed on a single-GPU. Benchmarks with parallelization may be parallelized across GPUs on 1-Node or across 2-Nodes, and may imply data parallelization or model parallelization. Libraries are: \includegraphics[height=2ex]{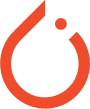} Pytorch, \includegraphics[height=2ex]{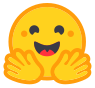} Huggingface, \includegraphics[height=2ex]{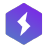} Pytorch-lightning, \includegraphics[height=2ex]{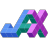} JAX and \includegraphics[height=2ex]{img/pyg\_logo.png} Pytorch-geometric. Unit of work is the element we use for the benchmarks performance metric, e.g., the number of sentences processed per second.}
\begin{tabular}{lrl|lll|llll}
                                                 &                                   &                                    & \multicolumn{3}{c|}{\textbf{Model}}                                                                           &                                                                                                       &                                          &                                  &                                         \\
\textbf{Bench}                 & \textbf{Weight} & \textbf{Domains} & \textbf{Architecture}              & \textbf{Name}                             & \textbf{Size}               & \textbf{Dataset/Env.}                                                               & \textbf{Parallelism}   & \textbf{Libs}  & \textbf{Unit of work} \\ \hline
\rowcolor[HTML]{FFFFFF} 
reformer                         & 1                                                     & NLP                                & Transformer           & Reformer           & 6M                                           & Synthetic                                                                      &                                        & \includegraphics[height=2ex]{img/pytorch-logo.png} \includegraphics[height=2ex]{img/huggingface-logo.png}                       & Sentences                                \\
\rowcolor[HTML]{F3F3F3} 
bert-tf32-fp16                   & 1                                                     & NLP                                & Transformer           & BERT               & 116M                                         & Synthetic                                                                      &                                        & \includegraphics[height=2ex]{img/pytorch-logo.png} \includegraphics[height=2ex]{img/huggingface-logo.png}                       & Sentences                                \\
\rowcolor[HTML]{FFFFFF} 
llm-lora-single                  & 1                                                     & NLP                                & Transformer           & Llama 3.2 70B      & 70B                                          & alpaca                                                                         &                                        & \includegraphics[height=2ex]{img/pytorch-logo.png}         \includegraphics[height=2ex]{img/huggingface-logo.png}         & Tokens                                  \\
\rowcolor[HTML]{F3F3F3} 
llm-lora-ddp-gpus                & 1                                                     & NLP                                & Transformer           & Llama 3.2 70B      & 70B                                          & alpaca                                                                         & 1-Node Data-Par.                       & \includegraphics[height=2ex]{img/pytorch-logo.png} \includegraphics[height=2ex]{img/huggingface-logo.png}                       & Tokens                                  \\
\rowcolor[HTML]{FFFFFF} 
llm-lora-ddp-nodes               & 1                                                     & NLP                                & Transformer           & Llama 3.2 70B      & 70B                                          & alpaca                                                                         & 2-Nodes Data-Par.                      & \includegraphics[height=2ex]{img/pytorch-logo.png} \includegraphics[height=2ex]{img/huggingface-logo.png}                       & Tokens                                  \\
\rowcolor[HTML]{F3F3F3} 
llm-lora-mp-gpus                 & 1                                                     & NLP                                & Transformer           & Llama 3.2 70B      & 70B                                          & alpaca                                                                         & 1-Node Model-Par.                      & \includegraphics[height=2ex]{img/pytorch-logo.png} \includegraphics[height=2ex]{img/huggingface-logo.png}                       & Tokens                                  \\

\rowcolor[HTML]{F3F3F3} 
llm-full-mp-nodes                & 1                                                     & NLP                                & Transformer           & Llama 3.2 70B      & 70B                                          & alpaca                                                                         & 2-Nodes Model-Par.                     & \includegraphics[height=2ex]{img/pytorch-logo.png} \includegraphics[height=2ex]{img/huggingface-logo.png}                       & Tokens                                  \\
\rowcolor[HTML]{FFFFFF} 
llama                            & 1                                                     & NLP                                & Transformer           & Llama 2.0 7B       & 7B                                           & wikitext-103-v1                                                                &                                        & \includegraphics[height=2ex]{img/pytorch-logo.png} \includegraphics[height=2ex]{img/huggingface-logo.png}                       & Tokens                                  \\
\rowcolor[HTML]{F3F3F3} 
vjepa-single                     & 1                                                     & CV                                 & Transformer           & V-JEPA             & 632M                                         & FakeVideos                                                                     &                                        & \includegraphics[height=2ex]{img/pytorch-logo.png}                            & Images                                  \\
\rowcolor[HTML]{FFFFFF} 
vjepa-gpus                       & 1                                                     & CV                                 & Transformer           & V-JEPA             & 632M                                         & FakeVideos                                                                     & 1-Node Data-Par.                       & \includegraphics[height=2ex]{img/pytorch-logo.png}                            & Images                                  \\
\rowcolor[HTML]{F3F3F3} 
resnet50                         & 1                                                     & CV                                 & CNN                   & ResNet50           & 26M                                          & FakeImageNet                                                                   &                                        & \includegraphics[height=2ex]{img/pytorch-logo.png}                            & Images                                  \\
\rowcolor[HTML]{FFFFFF} 
lightning-gpus                   & 1                                                     & CV                                 & CNN                   & ResNet152          & 60M                                          & FakeImageNet                                                                   & 1-Node Data-Par.                       & \includegraphics[height=2ex]{img/pytorch-logo.png} \includegraphics[height=2ex]{img/pytorch-lightning-logo.png}                       & Images                                  \\
\rowcolor[HTML]{F3F3F3} 
convnext\_large-tf32-fp16        & 1                                                     & CV                                 & CNN                   & ConvNext Large     & 200M                                         & FakeImageNet                                                                   &                                        & \includegraphics[height=2ex]{img/pytorch-logo.png}                            & Images                                  \\
\rowcolor[HTML]{FFFFFF} 
regnet\_y\_128gf                 & 1                                                     & CV                                 & CNN, RNN              & RegNet Y 128GF     & 693M                                         & FakeImageNet                                                                   &                                        & \includegraphics[height=2ex]{img/pytorch-logo.png}                            & Images                                  \\
\rowcolor[HTML]{F3F3F3} 
dinov2-giant-gpus                & 1                                                     & CV                                 & Transformer           & ViT-g/14           & 1B                                           & FakeImageNet                                                                   & 1-Node Data-Par.                       & \includegraphics[height=2ex]{img/pytorch-logo.png}                            & Images                                  \\
\rowcolor[HTML]{FFFFFF} 
diffusion-gpus                   & 1                                                     & CV, NLP                            & Transformer           & stable-diffusion-2 & 1B                   & naruto-blip-captions                                                           & 1-Node Data-Par.                       & \includegraphics[height=2ex]{img/pytorch-logo.png} \includegraphics[height=2ex]{img/huggingface-logo.png}                       & Images                                  \\
\rowcolor[HTML]{F3F3F3} 
diffusion-nodes                  & 1                                                     & CV, NLP                            & Transformer           & stable-diffusion-2 & 1B                   & naruto-blip-captions                                                           & 2-Nodes Data-Par.                      & \includegraphics[height=2ex]{img/pytorch-logo.png} \includegraphics[height=2ex]{img/huggingface-logo.png}                       & Images                                  \\
\rowcolor[HTML]{FFFFFF} 
llava-single                     & 1                                                     & CV, NLP                            & Transformer           & llava-1.5-7b-hf    & 7B                                           & The Cauldron        &                                        & \includegraphics[height=2ex]{img/pytorch-logo.png} \includegraphics[height=2ex]{img/huggingface-logo.png}                       & Images                                  \\
\rowcolor[HTML]{F3F3F3} 
torchatari                       & 1                                                     & CV, RL                             & CNN                   & -                  & 1.7M & Breakout-v5                                                                    &                                        & \includegraphics[height=2ex]{img/pytorch-logo.png}                            & Steps                                   \\
\rowcolor[HTML]{FFFFFF} 
ppo                              & 1                                                     & RL                             & MLP & -                  & 139K & hopper                                                                         &                                        & \includegraphics[height=2ex]{img/jax-logo.png}                           & Steps                                   \\
\rowcolor[HTML]{F3F3F3} 
brax                             & 1                                                     & RL                                 & MLP                   & -                  & 275K & ant                                                                            &                                        & \includegraphics[height=2ex]{img/jax-logo.png}                           & Steps                                   \\
\rowcolor[HTML]{FFFFFF} 
rlhf-single                      & 1                                                     & NLP, RL                            & Transformer           & pythia-1b-deduped  & 7B                                           & \begin{tabular}[c]{@{}l@{}}descriptiveness-\\ sentiment-trl-style\end{tabular} &                                        & \includegraphics[height=2ex]{img/pytorch-logo.png} \includegraphics[height=2ex]{img/huggingface-logo.png}                       & Tokens                                  \\
\rowcolor[HTML]{F3F3F3} 
pna                              & 2                                                     & Graphs                             & GNN                   & PNA                & 4M & PCQM4Mv2                                                                       &                                        & \includegraphics[height=2ex]{img/pytorch-logo.png} \includegraphics[height=2ex]{img/pyg\_logo.png}                      & Molecules                               \\
\rowcolor[HTML]{FFFFFF} 
dimenet                          & 2                                                     & Graphs                             & GNN                   & DimeNet            & 500K & PCQM4Mv2                                                                       &                                        & \includegraphics[height=2ex]{img/pytorch-logo.png} \includegraphics[height=2ex]{img/pyg\_logo.png}                      & Molecules                               \\
\rowcolor[HTML]{F3F3F3} 
recursiongfn                     & 2                                                     & Graphs                             & GFlowNet, T.          & GFlowNet           & 600M & N/A                                                                            &                                        & \includegraphics[height=2ex]{img/pytorch-logo.png} \includegraphics[height=2ex]{img/pyg\_logo.png}                      & Graph nodes                            
\end{tabular}
\label{tab:main-benchmarks}
\egroup
\end{table}

\begin{table}[ht]
\bgroup
\def\arraystretch{1.2}
\vspace{-4em}
\caption{Additional benchmarks included in Milabench.  All additional benchmarks are variations on main benchmarks except for
the 4 benchmarks at the top of the table. The main benchmarks are included
in pale gray along with their variations in dark. The 4 first benchmarks provide 
raw flops performance over precision formats, while the variations on \texttt{convnext\_large} and
\texttt{bert} provide a measure of performance over precision formats on 
standard training pipelines. The variations on \texttt{resnet50} allows
measuring the impact of I/O on the performance measure.
Finally, the variations on \texttt{lightning}, \texttt{dinov2}, 
\texttt{diffusion} and \texttt{rlhf} provides additional scaling measures, from 
single-GPU execution up to multi-node parallelization.
See Table~\ref{tab:main-benchmarks} for more information on the columns.}
\begin{tabular}{lrl|lll|llll}
                                                 &                                   &                                    & \multicolumn{3}{c|}{\textbf{Model}}                                                                           &                                                                                                       &                                          &                                  &                                         \\
\textbf{Bench}                 & \textbf{Weight} & \textbf{Domains} & \textbf{Architecture}              & \textbf{Name}                             & \textbf{Size}               & \textbf{Dataset/Env.}                                                               & \textbf{Parallelism}   & \textbf{Libs}  & \textbf{Unit of work} \\ \hline
\rowcolor[HTML]{FFFFFF} 
bf16                                             & 0                                                     & N/A                                & N/A                                & N/A                                       & N/A                         & N/A                                                                                                   &                                          & \includegraphics[height=2ex]{img/pytorch-logo.png}                             & Float Op.                \\
\rowcolor[HTML]{F3F3F3} 
fp16                                             & 0                                                     & N/A                                & N/A                                & N/A                                       & N/A                         & N/A                                                                                                   &                                          & \includegraphics[height=2ex]{img/pytorch-logo.png}                             & Float Op.                \\
\rowcolor[HTML]{FFFFFF} 
tf32                                             & 0                                                     & N/A                                & N/A                                & N/A                                       & N/A                         & N/A                                                                                                   &                                          & \includegraphics[height=2ex]{img/pytorch-logo.png}                             & Float Op.                \\
\rowcolor[HTML]{F3F3F3} 
fp32                                             & 0                                                     & N/A                                & N/A                                & N/A                                       & N/A                         & N/A                                                                                                   &                                          & \includegraphics[height=2ex]{img/pytorch-logo.png}                             & Float Op.                \\
\rowcolor[HTML]{FFFFFF} 
{\color[HTML]{CCCCCC} convnext\_large-tf32-fp16} & {\color[HTML]{CCCCCC} 1}                              & {\color[HTML]{CCCCCC} CV}          & {\color[HTML]{CCCCCC} CNN}         & {\color[HTML]{CCCCCC} ConvNext Large}     & {\color[HTML]{CCCCCC} 200M} & {\color[HTML]{CCCCCC} FakeImageNet}                                                                   &                                          & {\color[HTML]{CCCCCC} \includegraphics[height=2ex]{img/pytorch-logo.png}}      & {\color[HTML]{CCCCCC} Images}           \\
\rowcolor[HTML]{F3F3F3} 
convnext\_large-fp16                             & 0                                                     & CV                                 & CNN                                & ConvNext Large                            & 200M                        & FakeImageNet                                                                                          &                                          & \includegraphics[height=2ex]{img/pytorch-logo.png}                             & Images                                  \\
\rowcolor[HTML]{FFFFFF} 
convnext\_large-tf32                             & 0                                                     & CV                                 & CNN                                & ConvNext Large                            & 200M                        & FakeImageNet                                                                                          &                                          & \includegraphics[height=2ex]{img/pytorch-logo.png}                             & Images                                  \\
\rowcolor[HTML]{F3F3F3} 
convnext\_large-fp32                             & 0                                                     & CV                                 & CNN                                & ConvNext Large                            & 200M                        & FakeImageNet                                                                                          &                                          & \includegraphics[height=2ex]{img/pytorch-logo.png}                             & Images                                  \\
\rowcolor[HTML]{FFFFFF} 
{\color[HTML]{CCCCCC} bert-tf32-fp16}            & {\color[HTML]{CCCCCC} 1}                              & {\color[HTML]{CCCCCC} NLP}         & {\color[HTML]{CCCCCC} Transformer} & {\color[HTML]{CCCCCC} BERT}               & {\color[HTML]{CCCCCC} 116M} & {\color[HTML]{CCCCCC} Synthetic}                                                                      &                                          & {\color[HTML]{CCCCCC} \includegraphics[height=2ex]{img/pytorch-logo.png} \includegraphics[height=2ex]{img/huggingface-logo.png}} & {\color[HTML]{CCCCCC} Sentences}         \\
\rowcolor[HTML]{F3F3F3} 
bert-fp16                                        & 0                                                     & NLP                                & Transformer                        & BERT                                      & 116M                        & Synthetic                                                                                             &                                          & \includegraphics[height=2ex]{img/pytorch-logo.png} \includegraphics[height=2ex]{img/huggingface-logo.png}                        & Sentences                                \\
\rowcolor[HTML]{FFFFFF} 
bert-tf32                                        & 0                                                     & NLP                                & Transformer                        & BERT                                      & 116M                        & Synthetic                                                                                             &                                          & \includegraphics[height=2ex]{img/pytorch-logo.png} \includegraphics[height=2ex]{img/huggingface-logo.png}                        & Sentences                                \\
\rowcolor[HTML]{F3F3F3} 
bert-fp32                                        & 0                                                     & NLP                                & Transformer                        & BERT                                      & 116M                        & Synthetic                                                                                             &                                          & \includegraphics[height=2ex]{img/pytorch-logo.png} \includegraphics[height=2ex]{img/huggingface-logo.png}                        & Sentences                                \\ \hline
\rowcolor[HTML]{FFFFFF} 
{\color[HTML]{CCCCCC} resnet50}                  & {\color[HTML]{CCCCCC} 1}                              & {\color[HTML]{CCCCCC} CV}          & {\color[HTML]{CCCCCC} CNN}         & {\color[HTML]{CCCCCC} ResNet50}           & {\color[HTML]{CCCCCC} 26M}  & {\color[HTML]{CCCCCC} FakeImageNet}                                                                   &                                          & {\color[HTML]{CCCCCC} \includegraphics[height=2ex]{img/pytorch-logo.png}}      & {\color[HTML]{CCCCCC} Images}           \\
\rowcolor[HTML]{F3F3F3} 
resnet50-noio                                    & 0                                                     & CV                                 & CNN                                & ResNet50                                  & 26M                         & FakeImageNet                                                                                          &                                          & \includegraphics[height=2ex]{img/pytorch-logo.png}                             & Images                                  \\ \hline
\rowcolor[HTML]{FFFFFF} 
lightning                                        & 0                                                     & CV                                 & CNN                                & ResNet152                                 & 60M                         & FakeImageNet                                                                                          &                                          & \includegraphics[height=2ex]{img/pytorch-logo.png} \includegraphics[height=2ex]{img/pytorch-lightning-logo.png}                        & Images                                  \\
\rowcolor[HTML]{F3F3F3} 
{\color[HTML]{CCCCCC} lightning-gpus}            & {\color[HTML]{CCCCCC} 1}                              & {\color[HTML]{CCCCCC} CV}          & {\color[HTML]{CCCCCC} CNN}         & {\color[HTML]{CCCCCC} ResNet152}          & {\color[HTML]{CCCCCC} 60M}  & {\color[HTML]{CCCCCC} FakeImageNet}                                                                   & {\color[HTML]{CCCCCC} 1-Node Data-Par.}  & {\color[HTML]{CCCCCC} \includegraphics[height=2ex]{img/pytorch-logo.png} \includegraphics[height=2ex]{img/pytorch-lightning-logo.png}} & {\color[HTML]{CCCCCC} Images}           \\
\rowcolor[HTML]{FFFFFF} 
dinov2-giant-single                              & 0                                                     & CV                                 & Transformer                        & ViT-g/14                                  & 1B                          & FakeImageNet                                                                                          &                                          & \includegraphics[height=2ex]{img/pytorch-logo.png}                             & Images                                  \\
\rowcolor[HTML]{F3F3F3} 
{\color[HTML]{CCCCCC} dinov2-giant-gpus}         & {\color[HTML]{CCCCCC} 1}                              & {\color[HTML]{CCCCCC} CV}          & {\color[HTML]{CCCCCC} Transformer} & {\color[HTML]{CCCCCC} ViT-g/14}           & {\color[HTML]{CCCCCC} 1B}   & {\color[HTML]{CCCCCC} FakeImageNet}                                                                   & {\color[HTML]{CCCCCC} 1-Node Data-Par.}  & {\color[HTML]{CCCCCC} \includegraphics[height=2ex]{img/pytorch-logo.png}}      & {\color[HTML]{CCCCCC} Images}           \\
\rowcolor[HTML]{FFFFFF} 
diffusion-single                                 & 0                                                     & CV, NLP                            & Transformer                        & stable-diffusion-2                        & 1B                          & naruto-blip-captions                                                                                  &                                          & \includegraphics[height=2ex]{img/pytorch-logo.png} \includegraphics[height=2ex]{img/huggingface-logo.png}                        & Images                                  \\
\rowcolor[HTML]{F3F3F3} 
{\color[HTML]{CCCCCC} diffusion-gpus}            & {\color[HTML]{CCCCCC} 1}                              & {\color[HTML]{CCCCCC} CV, NLP}     & {\color[HTML]{CCCCCC} Transformer} & {\color[HTML]{CCCCCC} stable-diffusion-2} & {\color[HTML]{CCCCCC} 1B}   & {\color[HTML]{CCCCCC} naruto-blip-captions}                                                           & {\color[HTML]{CCCCCC} 1-Node Data-Par.}  & {\color[HTML]{CCCCCC} \includegraphics[height=2ex]{img/pytorch-logo.png} \includegraphics[height=2ex]{img/huggingface-logo.png}} & {\color[HTML]{CCCCCC} Images}           \\
\rowcolor[HTML]{FFFFFF} 
{\color[HTML]{CCCCCC} diffusion-nodes}           & {\color[HTML]{CCCCCC} 1}                              & {\color[HTML]{CCCCCC} CV, NLP}     & {\color[HTML]{CCCCCC} Transformer} & {\color[HTML]{CCCCCC} stable-diffusion-2} & {\color[HTML]{CCCCCC} 1B}   & {\color[HTML]{CCCCCC} naruto-blip-captions}                                                           & {\color[HTML]{CCCCCC} 2-Nodes Data-Par.} & {\color[HTML]{CCCCCC} \includegraphics[height=2ex]{img/pytorch-logo.png} \includegraphics[height=2ex]{img/huggingface-logo.png}} & {\color[HTML]{CCCCCC} Images}           \\
\rowcolor[HTML]{F3F3F3} 
{\color[HTML]{CCCCCC} rlhf-single}               & {\color[HTML]{CCCCCC} 1}                              & {\color[HTML]{CCCCCC} NLP, RL}     & {\color[HTML]{CCCCCC} Transformer} & {\color[HTML]{CCCCCC} pythia-1b-deduped}  & {\color[HTML]{CCCCCC} 7B}   & {\color[HTML]{CCCCCC} \begin{tabular}[c]{@{}l@{}}descriptiveness-\\ sentiment-trl-style\end{tabular}} &                                          & {\color[HTML]{CCCCCC} \includegraphics[height=2ex]{img/pytorch-logo.png} \includegraphics[height=2ex]{img/huggingface-logo.png}} & {\color[HTML]{CCCCCC} Tokens}           \\
\rowcolor[HTML]{FFFFFF} 
rlhf-gpus                                        & 0                                                     & NLP, RL                            & Transformer                        & pythia-1b-deduped                         & 7B                          & \begin{tabular}[c]{@{}l@{}}descriptiveness-\\ sentiment-trl-style\end{tabular}                        & 1-Node Data-Par.                         & \includegraphics[height=2ex]{img/pytorch-logo.png} \includegraphics[height=2ex]{img/huggingface-logo.png}                        & Tokens                                 
\end{tabular}
\label{tab:additional-benchmarks}
\egroup
\end{table}

\end{landscape}

\begin{table}[ht]
\caption{Results on main benchmarks}
\bgroup
\def\arraystretch{1.2}
\begin{tabular}{l|rrr|rrrr}
                          & \multicolumn{3}{c}{\textbf{Ratio with A100}}                                                                & \multicolumn{4}{c}{\textbf{Performance}}                                                                                    \\
\textbf{Bench}            & \textbf{H100} & \textbf{MI300X}                              & \textbf{Gaudi2}                              & \textbf{A100} & \textbf{H100} & \textbf{MI300X}                              & \textbf{Gaudi2}                              \\ \hline
\rowcolor[HTML]{FFFFFF} 
reformer                  & 1.67          & 0.81                                         & 0.63                                         & 62.3          & 103.7         & 50.5                                         & 39.1                                         \\
\rowcolor[HTML]{F3F3F3} 
bert-tf32-fp16            & 1.75          & 0.77                                         & 0.66                                         & 264.7         & 462.9         & 202.6                                        & 175.8                                        \\
\rowcolor[HTML]{FFFFFF} 
llm-lora-single           & 1.86          & 0.98                                         & 0.41                                         & 2.7K          & 5.1K          & 2.7K                                         & 1.1K                                         \\
\rowcolor[HTML]{F3F3F3} 
llm-lora-ddp-gpus         & 1.75          & 0.79                                         & 0.32                                         & 16.8K         & 29.3K         & 13.2K                                        & 5.3K                                         \\
\rowcolor[HTML]{FFFFFF} 
llm-lora-ddp-nodes        & 3.13          & 1.78                                         &  & 17.9K         & 56.2K         & 31.9K                                        &  \\
\rowcolor[HTML]{F3F3F3} 
llm-lora-mp-gpus          & 1.91          & 1.11                                         & 0.34                                         & 2.0K          & 3.8K          & 2.2K                                         & 680.0                                        \\
\rowcolor[HTML]{FFFFFF} 
llm-full-mp-gpus          & 2.32          & 1.68                                         & 2.15                                         & 195.2         & 453.4         & 327.8                                        & 419.3                                        \\
\rowcolor[HTML]{F3F3F3} 
llm-full-mp-nodes         & 5.51          & 4.87                                         &  & 146.1         & 805.3         & 710.7                                        &  \\
\rowcolor[HTML]{FFFFFF} 
llama                     & 1.57          & 0.19                                         & 0.43                                         & 493.3         & 774.0         & 92.6                                         & 211.0                                        \\
\rowcolor[HTML]{F3F3F3} 
vjepa-gpus                & 2.03          & 0.58                                         & 0.31                                         & 127.7         & 259.8         & 73.6                                         & 40.1                                         \\
\rowcolor[HTML]{FFFFFF} 
vjepa-single              & 1.91          & 0.55                                         & 0.25                                         & 21.3          & 40.8          & 11.8                                         & 5.3                                          \\
\rowcolor[HTML]{F3F3F3} 
resnet50                  & 1.96          & 1.97                                         & 3.17                                         & 854.3         & 1.7K          & 1.7K                                         & 2.7K                                         \\
\rowcolor[HTML]{FFFFFF} 
lightning-gpus            & 3.09          & 1.86                                         & 1.10                                         & 3.1K          & 9.6K          & 5.8K                                         & 3.4K                                         \\
\rowcolor[HTML]{F3F3F3} 
convnext\_large-tf32-fp16 & 1.95          & 0.71                                         & 0.86                                         & 339.1         & 662.8         & 239.5                                        & 293.2                                        \\
\rowcolor[HTML]{FFFFFF} 
regnet\_y\_128gf          & 1.57          & 0.85                                         & 1.45                                         & 119.5         & 187.3         & 102.0                                        & 173.3                                        \\
\rowcolor[HTML]{F3F3F3} 
dinov2-giant-gpus         & 1.92          &  &  & 447.1         & 856.8         &  &  \\
\rowcolor[HTML]{FFFFFF} 
diffusion-gpus            & 3.16          & 0.94                                         &  & 120.3         & 380.1         & 113.5                                        &  \\
\rowcolor[HTML]{F3F3F3} 
diffusion-nodes           & 3.41          & 0.94                                         &  & 227.6         & 775.2         & 212.8                                        &  \\
\rowcolor[HTML]{FFFFFF} 
llava-single              & 1.75          & 0.88                                         & 0.31                                         & 2.3           & 4.0           & 2.0                                          & 0.7                                          \\
\rowcolor[HTML]{F3F3F3} 
torchatari                & 1.54          & 0.64                                         & 0.49                                         & 6.0K          & 9.3K          & 3.9K                                         & 3.0K                                         \\
\rowcolor[HTML]{FFFFFF} 
ppo                       & 1.20          & 0.67                                         &  & 32.2M         & 38.8M         & 21.5M                                        &  \\
\rowcolor[HTML]{F3F3F3} 
brax                      & 1.21          & 0.23                                         &  & 727.5K        & 877.9K        & 170.4K                                       &  \\
\rowcolor[HTML]{FFFFFF} 
rlhf-single               & 2.75          & 1.56                                         &  & 1.1K          & 3.1K          & 1.8K                                         &  \\
\rowcolor[HTML]{F3F3F3} 
pna                       & 1.66          & 0.67                                         &  & 4.0K          & 6.6K          & 2.7K                                         &  \\
\rowcolor[HTML]{FFFFFF} 
dimenet                   & 1.50          & 0.64                                         &  & 373.1         & 560.2         & 237.6                                        &  \\
\rowcolor[HTML]{F3F3F3} 
recursiongfn              & 1.47          & 1.21                                         &  & 7.4K          & 10.9K         & 8.9K                                         &  \\ \hline
Global Score              & 1.93          & 0.74                                         & 0.02                                         & 1170.9       & 2263.7       & 866.7                                       & 24.8                                       
\end{tabular}
\egroup
\label{tab:main-results}
\end{table}

\begin{table}[ht]
\caption{Results on optional benchmarks.}
\bgroup
\def\arraystretch{1.2}
\begin{tabular}{l|rrr|rrrr}
                                                 & \multicolumn{3}{c|}{\textbf{Ratio with A100}}                                                                              & \multicolumn{4}{c}{\textbf{Performance}}                                                                                                              \\
\textbf{Bench}                                   & \textbf{H100}               & \textbf{MI300X}                              & \textbf{Gaudi2}                              & \textbf{A100}              & \textbf{H100}              & \textbf{MI300X}                              & \textbf{Gaudi2}                              \\ \hline
\rowcolor[HTML]{FFFFFF} 
bf16                                             & 2.67                        & 2.65                                         & 1.44                                         & 293                        & 784                        & 777                                          & 422                                          \\
\rowcolor[HTML]{F3F3F3} 
fp16                                             & 2.76                        & 2.67                                         & 1.47                                         & 289                        & 797                        & 772                                          & 427                                          \\
\rowcolor[HTML]{FFFFFF} 
tf32                                             & 2.82                        & 0.76                                         & 0.73                                         & 146                        & 413                        & 111                                          & 107                                          \\
\rowcolor[HTML]{F3F3F3} 
fp32                                             & 2.71                        & 5.78                                         & 5.60                                         & 19                         & 52                         & 111                                          & 107                                          \\
\rowcolor[HTML]{FFFFFF} 
{\color[HTML]{CCCCCC} convnext\_large-tf32-fp16} & {\color[HTML]{CCCCCC} 1.95} & {\color[HTML]{CCCCCC} 0.71}                  & {\color[HTML]{CCCCCC} 0.86}                  & {\color[HTML]{CCCCCC} 339} & {\color[HTML]{CCCCCC} 663} & {\color[HTML]{CCCCCC} 240}                   & {\color[HTML]{CCCCCC} 293}                   \\
\rowcolor[HTML]{F3F3F3} 
convnext\_large-fp16                             & 1.98                        & 0.72                                         & 0.88                                         & 334                        & 659                        & 240                                          & 293                                          \\
\rowcolor[HTML]{FFFFFF} 
convnext\_large-tf32                             & 1.54                        & 1.22                                         & 1.01                                         & 156                        & 239                        & 189                                          & 157                                          \\
\rowcolor[HTML]{F3F3F3} 
convnext\_large-fp32                             & 2.17                        & 3.20                                         & 2.65                                         & 60                         & 129                        & 190                                          & 157                                          \\
\rowcolor[HTML]{FFFFFF} 
{\color[HTML]{CCCCCC} bert-tf32-fp16}            & {\color[HTML]{CCCCCC} 1.75} & {\color[HTML]{CCCCCC} 0.77}                  & {\color[HTML]{CCCCCC} 0.66}                  & {\color[HTML]{CCCCCC} 265} & {\color[HTML]{CCCCCC} 463} & {\color[HTML]{CCCCCC} 203}                   & {\color[HTML]{CCCCCC} 176}                   \\
\rowcolor[HTML]{F3F3F3} 
bert-fp16                                        & 1.75                        & 0.77                                         & 0.65                                         & 265                        & 462                        & 203                                          & 172                                          \\
\rowcolor[HTML]{FFFFFF} 
bert-tf32                                        & 1.72                        & 0.52                                         & 0.90                                         & 142                        & 244                        & 74                                           & 128                                          \\
\rowcolor[HTML]{F3F3F3} 
bert-fp32                                        & 2.48                        & 1.65                                         & 2.86                                         & 45                         & 111                        & 74                                           & 128                                          \\ \hline
\rowcolor[HTML]{FFFFFF} 
{\color[HTML]{CCCCCC} resnet50}                  & {\color[HTML]{CCCCCC} 1.96} & {\color[HTML]{CCCCCC} 1.97}                  & {\color[HTML]{CCCCCC} 3.17}                  & {\color[HTML]{CCCCCC} 854} & {\color[HTML]{CCCCCC} 1K}  & {\color[HTML]{CCCCCC} 1K}                    & {\color[HTML]{CCCCCC} 2K}                    \\
\rowcolor[HTML]{F3F3F3} 
resnet50-noio                                    & 1.77                        & 1.91                                         & 3.47                                         & 1K                         & 2K                         & 2K                                           & 4K                                           \\ \hline
\rowcolor[HTML]{FFFFFF} 
lightning                                        & 1.80                        & 1.49                                         & 1.60                                         & 681                        & 1K                         & 1K                                           & 1K                                           \\
\rowcolor[HTML]{F3F3F3} 
{\color[HTML]{CCCCCC} lightning-gpus}            & {\color[HTML]{CCCCCC} 3.09} & {\color[HTML]{CCCCCC} 1.86}                  & {\color[HTML]{CCCCCC} 1.10}                  & {\color[HTML]{CCCCCC} 3K}  & {\color[HTML]{CCCCCC} 9K}  & {\color[HTML]{CCCCCC} 5K}                    & {\color[HTML]{CCCCCC} 3K}                    \\
\rowcolor[HTML]{FFFFFF} 
dinov2-giant-single                              & 1.92                        &  &  & 54                         & 103                        &  &  \\
\rowcolor[HTML]{F3F3F3} 
{\color[HTML]{CCCCCC} dinov2-giant-gpus}         & {\color[HTML]{CCCCCC} 1.92} &  &  & {\color[HTML]{CCCCCC} 447} & {\color[HTML]{CCCCCC} 857} &  &  \\
\rowcolor[HTML]{FFFFFF} 
diffusion-single                                 & 2.11                        & 0.64                                         &  & 24                         & 51                         & 16                                           &  \\
\rowcolor[HTML]{F3F3F3} 
{\color[HTML]{CCCCCC} diffusion-gpus}            & {\color[HTML]{CCCCCC} 3.16} & {\color[HTML]{CCCCCC} 0.94}                  &  & {\color[HTML]{CCCCCC} 120} & {\color[HTML]{CCCCCC} 380} & {\color[HTML]{CCCCCC} 114}                   &  \\
\rowcolor[HTML]{FFFFFF} 
{\color[HTML]{CCCCCC} diffusion-nodes}           & {\color[HTML]{CCCCCC} 3.41} & {\color[HTML]{CCCCCC} 0.94}                  &  & {\color[HTML]{CCCCCC} 228} & {\color[HTML]{CCCCCC} 775} & {\color[HTML]{CCCCCC} 213}                   &  \\
\rowcolor[HTML]{F3F3F3} 
{\color[HTML]{CCCCCC} rlhf-single}               & {\color[HTML]{CCCCCC} 2.75} & {\color[HTML]{CCCCCC} 1.56}                  &  & {\color[HTML]{CCCCCC} 1K}  & {\color[HTML]{CCCCCC} 3K}  & {\color[HTML]{CCCCCC} 1K}                    &  \\
\rowcolor[HTML]{FFFFFF} 
rlhf-gpus                                        & 2.60                        & 1.98                                         &  & 6K                         & 16K                        & 12K                                          & 
\end{tabular}
\egroup
\label{tab:optional-results}
\end{table}

\clearpage

\appendix
\section{Hardware \& Configuration}


\newcommand\xrowht[2][0]{\addstackgap[.5\dimexpr#2\relax]{\vphantom{#1}}}

\begin{table}[ht]
\begin{adjustwidth}{-2cm}{}
\caption{Node specifications used for the results}
\begin{tabular}{ |l||l|l|l|l| }
\hline\xrowht{14pt}
         & \textbf{A100}                 &\textbf{H100}                  & \textbf{MI300X}                & \textbf{Gaudi2}                \\ 
\hline
\hline \xrowht{14pt}
\textbf{Provider}      &  Nvidia (Donated) & \multicolumn{2}{c|}{Dell (Remote access)} & Intel (Remote access)  \\ \hline
\xrowht{14pt}
\textbf{System}      &  DGX A100  & \multicolumn{2}{c|}{Dell PowerEdge XE9680} & Supermicro X12DPG-OA6-GD2  \\
\hline\xrowht{14pt}
\textbf{OS}      &  \multicolumn{4}{c|}{Ubuntu 22.04}   \\
\hline\xrowht{14pt}
\textbf{CPU}      &  2x AMD EPYC 7742  & \multicolumn{2}{c|}{2x Intel Xeon Platinum 8468 } & 2x Intel Xeon Platinum 8380 CPU   \\
\hline\xrowht{14pt}
\textbf{CPU Cores} & 128                  &  \multicolumn{2}{c|}{ 96 }   &   80                                          \\
\hline\xrowht{14pt}
\textbf{RAM}      & 2 TB                 & 1 TB  & 2 TB                 & 1 TB                  \\
\hline\xrowht{14pt}
\textbf{GPU}      & 8x A100 SXM4 80Go   & 8x H100 SXM5 80Go    & 8x MI300X 192 GB    & 8x Gaudi2 96 GB      \\
\hline\xrowht{14pt}
\textbf{Pytorch}      & 2.4.0   & 2.4.0    & 2.4.1       & 2.3      \\
\hline\xrowht{14pt}
\textbf{Version}      & cuda 12.1   &  cuda 12.1   & ROCm 6.1    & 1.17      \\
\hline\xrowht{14pt}
\textbf{Network}  & MT28908 [ConnectX-6] & \multicolumn{2}{c|}{ Mellanox ConnectX-6 }                  & N/A                  \\
\hline\xrowht{14pt}
\textbf{Code}   & \multicolumn{4}{c|}{https://github.com/mila-iqia/milabench}                                   \\
\textbf{Branch}   & H100               &  H100                 & rocm                 & hpu                   \\
\hline
\end{tabular}
\label{tab:node-specs}
 \end{adjustwidth}
\end{table}

Note that due to timeline issues, not all vendors could be ran with the same version of PyTorch.
We use PyTorch 2.4 version or the latest available at the time. PyTorch version can have a major impact on performance.

\section{GPT-4o Response Example}
\label{app:gpt-4o}

\begin{verbatim}
"title": {
  "value": "UNMASKING THE LOTTERY TICKET HYPOTHESIS: WHAT’S ENCODED IN A
  WINNING TICKET’S MASK?",
  "justification": "This is the exact title of the paper provided in the
  given text.",
  "quote": "UNMASKING THE LOTTERY TICKET HYPOTHESIS: WHAT’S ENCODED IN A
  WINNING TICKET’S MASK?"
},
"primary_research_field": {
  "name": {
    "value": "Model Pruning",
    "justification": "The primary focus of the paper is on understanding
    iterative magnitude pruning (IMP) and its effectiveness in identifying
    sparse subnetworks that can be trained to achieve the same accuracy as
    the full network.",
    "quote": "A state of the art algorithm for doing so is iterative
    magnitude pruning (IMP) (Frankle et al., 2020a)."
  }
},
"sub_research_fields": [
  {
    "name": {
      "value": "Neural Network Training",
      "justification": "The research extensively analyzes how training
      procedures, such as gradient descent, interact with pruned networks
      and the robustness of these methods.",
      "quote": "Second, how does SGD starting from the masked rewind point
      extract and use this information?"
    }
  }
],
"models": [
  {
    "name": {
      "value": "ResNet-20",
      "justification": "ResNet-20 is one of the network architectures used
      for empirical investigations in this paper.",
      "quote": "We do so through extensive empirical investigations on
      [...] modern network architectures (ResNet-20, ResNet-18, and
      ResNet-50)."
    },
  },
  {
    "name": {
      "value": "ResNet-18",
      "justification": "ResNet-18 is one of the network architectures used
      for empirical investigations in this paper.",
      "quote": "We do so through extensive empirical investigations on
      [...] modern network architectures (ResNet-20, ResNet-18, and
      ResNet-50)."
    }
  },
  {
    "name": {
      "value": "ResNet-50",
      "justification": "ResNet-50 is one of the network architectures used
      for empirical investigations in this paper.",
      "quote": "We do so through extensive empirical investigations on
      [...] modern network architectures (ResNet-20, ResNet-18, and
      ResNet-50)."
    }
  }
]
\end{verbatim}

\clearpage
\section{Methodology}

Milabench use hardware event as timer to ensure we time the execution of the kernels in its entirety.
To avoid unnecessary sync, the events are appended during training and only logged at the end of the epochs.

To simplify the addition of new benchmarks, an iterator wrapper \texttt{TimedIterator} is provided, which 
encapsulate the necessary logic.

\lstdefinestyle{python}{language=Python,
    morekeywords={Event},
    morekeywords={StopProgram},
}

\begin{lstlisting}[style = python]
for i in range(epoch):
  events = []
  
  # Creation of the iterator from the dataloader is time consuming
  # it would get amortized across many batch during real training
  batch_iter = iter(loader)
  total_obs = 0
  
  # Avoid sync in the batch loop
  start = Event()
  start.record()
  
  for batch in batch_iter:
    
    pred = model(batch)
    loss = fn(pred, target)
                                                    
    end = Event()                                   # <-+ Limited
    end.record()                                    #   | overhead
    obs = (start, end, len(batch), loss.detach())   #   | happens 
    events.append(obs)                              #   | during  
    if len(events) + total_obs >= 60:               #   | kernel
      break                                         #   | runs
    start = end                                     # <-+

  # Force sync at the end of the epoch     # <-+
  for start, end, bs, loss  in events:     #   | Timer is off 
    end.wait()                             #   | logging does not
    log(loss=loss.item())                  #   | impact  
    log(rate=bs / (end - start))           #   | perf 
  total_obs += len(events)                 #   | measures
  if total_obs >= 60:                      #   |
    raise StopProgram()                    # <-+
\end{lstlisting}

Additionally, most of milabench benchmarks are research repository, to avoid having to fork or modify the original repository
milabench use libraries to hot patch the code where necessary to use milabench tools to record and measure performance.
This allows milabench to dynamically wrap dataloaders with its own \texttt{TimedIterator} without modifying the original code.

This simplifies the update of benchmarks (as we can simply pull the changes from the original repo)
and ensures the benchmark is as close as possible to the real world use case.

\end{document}